\documentclass[runningheads]{llncs}
\usepackage[T1]{fontenc}
\usepackage{graphicx}
\usepackage{hyperref}
\usepackage{xcolor}
\usepackage{amsmath}
\usepackage{amsfonts}
\usepackage{relsize}
\usepackage{booktabs}
\usepackage{colortbl}
\usepackage{placeins}
\usepackage{bold-extra}
\newcommand{\ind}{\hspace*{1cm}}

\newcommand{\aucmean}{\text{UER-AUC}_{\mathrm{mean}}}
\newcommand{\aucmax}{\text{UER-AUC}_{\mathrm{max}}}

\newcommand{\etal}{\textit{et al. }}

\begin{document}

\title{Improving Counterfactual Truthfulness for Molecular Property Prediction through Uncertainty Quantification}
\titlerunning{Uncertainty Quantification for Counterfactual Truthfulness}
\author{
Jonas Teufel\inst{1}\orcidID{0000-0002-9228-9395} \and
Annika Leinweber\inst{1} \and \\
Pascal Friederich\inst{1}$^{,}$\inst{2}\orcidID{0000-0003-4465-1465} 
}
\authorrunning{J. Teufel et al.}
\institute{ %
Institute of Theoretical Informatics, Karlsruhe Institute of Technology,\\
Kaiserstr. 12, 76131 Karlsruhe, Germany\\
\email{jonas.teufel@kit.edu}%, \email{annika.leinweber@student.kit.edu}
\and
Institute of Nanotechnology, Karlsruhe Institute of Technology,\\
Kaiserstr. 12, 76131 Karlsruhe, Germany\\
\email{pascal.friederich@kit.edu}
}
\maketitle              % typeset the header of the contribution
\begin{abstract}
Explainable AI (xAI) interventions aim to improve interpretability for complex black-box modes, not only to improve user trust but also as a means to extract scientific insights from high-performing predictive systems. In molecular property prediction, counterfactual explanations offer a way to understand predictive behavior by highlighting which minimal perturbations in the input molecular structure cause the greatest deviation in the predicted property. However, such explanations only allow for meaningful scientific insights if they reflect the distribution of the true underlying property---a feature we define as counterfactual truthfulness. To enhance truthfulness, we propose the integration of uncertainty estimation techniques to filter high-uncertainty counterfactuals. Through computational experiments with synthetic and real-world datasets, we demonstrate that combining traditional deep ensembles and mean variance estimation can substantially reduce average and maximum model error for out-of-distribution settings and especially increase counterfactual truthfulness. Our results highlight the importance of incorporating uncertainty estimation into counterfactual explainability, especially considering the relative effectiveness of low-effort strategies such as model ensembles.
\end{abstract}
% Corresponding author footnote
% \renewcommand{\thefootnote}{$\dagger$}
% \footnotetext{Corresponding author.}
% \renewcommand{\thefootnote}{\arabic{footnote}}
%
%
\section{Introduction}

% FIGURE: VISUAL ABSTRACT
% =======================
\begin{figure}[t]
    \centering
    \includegraphics[width=1.0\linewidth]{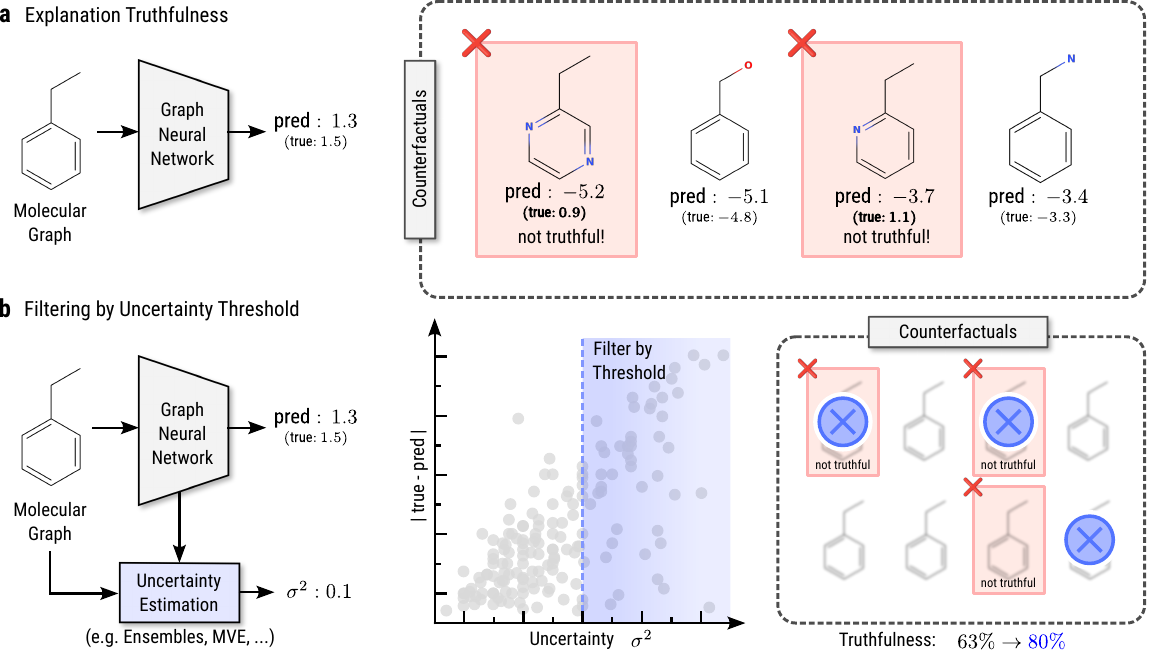}
    \caption{\textbf{\textsf{a}} Truthful explanations should not only reflect the model's behavior but also the properties of the underlying true data distribution. \textbf{\textsf{b}} Uncertainty quantification methods predict an additional uncertainty value as an approximation of the model's prediction error. By filtering high-uncertainty elements, it is possible to reduce the cumulative error and, by extension, the fraction of truthful counterfactuals of the remaining set.}
    \label{fig:abstract}
\end{figure}
% general introduction to xAI
Recent advances in the study of artificial intelligence (AI) have revolutionized various branches of society, industry, and science. Despite their numerous advantages, the opaque black-box nature of modern AI methods remains a challenge. Although complex neural network models often display superior predictive performance, their inner workings remain largely intransparent to humans. Explainable AI (xAI) aims to address these shortcomings by developing methods to better understand the inner workings of these complex models.

% how xAI can be relevant to scientific discovery.
Traditionally, xAI methods are meant to improve trust in human-AI relationships, provide tools for model debugging, and ensure regulatory compliance \cite{doshi-velezRigorousScienceInterpretable2017a}. More recently, xAI has been proposed as a potential source of new scientific insight \cite{zednikScientificExplorationExplainable2022,m.maffettoneWhatMissingAutonomous2023,teufelMEGANMultiexplanationGraph2023a,teufelGlobalConceptExplanations2024a}. This potential of gaining new insights primarily concerns tasks about which little to no prior human knowledge exists. By elucidating the behavior of high-performing models in complex property prediction tasks, xAI can offer insights not only into the model's behavior but, by extension, into the underlying rules and relationships governing the data itself. However, to gain meaningful insights, the given explanations must be \textit{truthful} regarding this underlying data distribution. However, for these explanations to yield meaningful insights, they must accurately reflect the true data distribution. This imposes a more stringent requirement for explanations: they must be valid not only in terms of the model's behavior but also with respect to the predicted property itself.

% specifically for molecular property prediction.
%This potential for new scientific insight primarily exists for the model. \\
% focus on counterfactuals.
In this work, we explore counterfactual explainability within chemistry and materials science—a domain where insights derived from XAI would have a substantial impact to accelerate scientific discovery. In short, counterfactual explanations locally explain the model's behavior by constructing multiple "what if?" scenarios of minimally perturbed input configurations that cause large deviations in the model's prediction. By itself, a counterfactual only has to explain the model's behavior, regardless if that behavior reflects the underlying property, causing significant conceptual overlap between the counterfactuals and adversarial examples \cite{freieslebenIntriguingRelationCounterfactual2022}. As an extension, we define a \textit{truthful} counterfactual as one that satisfies constraints regarding the model and the underlying ground truth---causing a large deviation of the prediction while maintaining low prediction error (see Figure~\ref{fig:abstract}).

% summary of the paper
Given the general unavailability of ground truth labels for counterfactual samples, we propose uncertainty quantification methods as a means to approximate prediction error and ultimately improve overall counterfactual truthfulness by filtering high-uncertainty explanations. We empirically investigate various common methods of uncertainty quantification and find that an ensemble of mean-variance estimators (MVE) yields the greatest improvement of relative model error and can substantially improve counterfactual truthfulness. Qualitative results affirm these findings, showing that uncertainty-based filtering removes unlikely molecular configurations that lie outside the training distribution.
Our results underscore the potential benefits of integrating uncertainty estimation into explainability methods, such as counterfactual explanations.
% One popular category of explanations is that of \textit{counterfactuals}. In short, counterfactuals provide a local explanation of a model's predictive behavior by presenting multiple "what if?" scenarios. Given an original input sample, a counterfactual is a minimal perturbation of that original input that creates a large deviation in the model's predicted output---thereby illustrating the model's local decision boundaries. In the specific context of chemistry and material science, counterfactuals are presented as similar input molecules for which property predictions greatly deviate. This way, molecular counterfactuals can provide direct feedback to domain experts on how to best modify a molecule to achieve a different outcome. However, a model can ... \\
% % Summary of what we do in this paper
% This work introduces the notion of \textit{counterfactual truthfulness} to measure how well a given set of counterfactual explanations reflects the ground truth properties of the underlying task. We argue that only with a high degree of truthfulness, it is possible to reliably draw conclusions from counterfactual explanations about the properties of the underlying task.\\

\section{Related Work}

% =====================
% GRAPH COUNTERFACTUALS
% =====================
% Starting with a general introduction of counterfactual explanations in the first place
\subsubsection{Graph Counterfactual Explanations. } Insights from social science indicate that humans prefer explanations to be \textit{contrastive}---to explain why something happened \textit{instead of} something else \cite{millerExplanationArtificialIntelligence2019}. Counterfactuals aim to provide such contrastive explanations by constructing hypothetical "what if?" scenarios to show which small perturbations to a given input sample would have resulted in a significant deviation from the original prediction outcome.

While Verma \etal \cite{vermaCounterfactualExplanationsAlgorithmic2024} present an extensive general review on the topic of counterfactual explanations across different data modalities, Prado-Romero \etal \cite{prado-romeroSurveyGraphCounterfactual2024} explore specifically counterfactual explanations in the graph processing domain. The authors find that the existing approaches can be categorized by which kinds of perturbations to the input graph are considered.
% Con examples: Lucic (CF-GNNExplainer), Baja (RCExplainer), Tan (CF2)
Many existing methods create perturbations using masking strategies on the node-, edge- or feature-level in which masks are optimized to maximize output deviations \cite{lucicCFGNNExplainerCounterfactualExplanations2021a,bajajRobustCounterfactualExplanations2021,tanLearningEvaluatingGraph2022a}. 
% pro examples: Wellawatte (MACCS), Numeroso (MEG), Nguyen (MACDA)
However, masking-based strategies often yield uninformative explanations for molecular property prediction. In this context, it is more insightful to perturb the molecular graph by adding or removing bonds and atoms \cite{sturmMitigatingMolecularAggregation2023a}. Some authors successfully adopt such approaches for molecular property predictions  \cite{wellawatteModelAgnosticGeneration2022a,nguyenExplainingBlackBox2023,numerosoMEGGeneratingMolecular2021}. One particular difficulty for these kinds of approaches is the necessity of including domain knowledge to ensure that modifications result in a valid graph structure (e.g. chemically feasible molecules). In one example, Numeroso and Bacciu \cite{numerosoMEGGeneratingMolecular2021}, propose to train an external reinforcement learning agent to propose suitable graph modifications resulting in counterfactual candidates for molecular property predictions. In this case, the authors also introduce domain knowledge by restricting the action space of the agent to chemically feasible modifications.
%
% ==========================
% UNCERTAINTY QUANTIFICATION
% ==========================
% General introduction to uncertainty estimation
% maybe something about then something about aleatoric vs. epistemic? 
% motivation about why uncertainty estimation is important in machine learning in general and it's connection to explainable AI specifically
% Gawlikows, Abdar (Reviews)
\subsubsection{Uncertainty Quantification.} Predictive machine learning models often encounter uncertainty from various sources, including, for example, inherent measurement noise (aleatoric uncertainty) or regions of the input space insufficiently covered in the training distribution (epistemic uncertainty). Consequently, a model's predictions may be more accurate for some input samples than for others. Uncertainty quantification methods aim to measure this variability, identifying those samples that a model can predict with greater confidence \cite{gawlikowskiSurveyUncertaintyDeep2023,abdarReviewUncertaintyQuantification2021}.

Similarly to the broader field of xAI, one aim of uncertainty quantification is to improve user trust by indicating the reliability of a prediction \cite{seussBridgingGapExplainable2021}. Beyond uncertainty quantification for target predictions, Longo~\etal \cite{longoExplainableArtificialIntelligence2024} propose to introduce elements of uncertainty estimation on the explanation level as well.

% Nix (MVE), 
% laksminarayan (Deep ensembles)
% Tishby; Bishop; Goan and Fookes (BNNs)
% Newer methods: The framework of repulsive ensembles?
% Most relevant category: Existing overlap between the two fields "counterfactuals" and "uncertainty estimation"
Traditionally used methods for uncertainty quantification include the joint prediction of a distribution's mean and variance (MVE) \cite{nixEstimatingMeanVariance1994}, assessing the variance between the predictions of a Deep Ensemble \cite{lakshminarayananSimpleScalablePredictive2017} and using bayesian neural networks (BNNs) \cite{tishbyConsistentInferenceProbabilities1989,bishopBayesianNeuralNetworks1997,goanBayesianNeuralNetworks2020} which aim to directly predict an output distribution rather than individual values. More recent alternatives include stochastic weight averaging gaussians (SWAG) \cite{maddoxSimpleBaselineBayesian2019} and the idea of Repulsive Ensembles \cite{dangeloRepulsiveDeepEnsembles2021,trinhInputgradientSpaceParticle2023} as an extension to Deep Ensembles built on the general framework of particle based variational inference (ParVI) \cite{liuSteinVariationalGradient2016,liuUnderstandingAcceleratingParticleBased2018} introducing explicit member diversification.

% Investigations for molecular property prediction
% Hirschfeld
% Scalia
In the domain of molecular property prediction, Hirschfeld \etal \cite{hirschfeldUncertaintyQuantificationUsing2020} and Scalia \etal \cite{scaliaEvaluatingScalableUncertainty2020} independently investigate the performance of various traditional uncertainty quantification methods across many standard property prediction datasets. Busk \etal specifically investigate uncertainty quantification using an ensemble of graph neural networks \cite{buskCalibratedUncertaintyMolecular2021}.
%
% ==============================================
% UNCERTAINTY QUANTIFICATION AND COUNTERFACTUALS
% ==============================================
%
\subsubsection{Uncertainty Quantification and Counterfactuals. } Using xAI to gain new insights into the underlying properties of the data distribution requires the given explanations to be \textit{truthful} regarding the \textit{true} property values. In the same context, Freiesleben \cite{freieslebenIntriguingRelationCounterfactual2022} addresses the conceptual distinction between counterfactual explanations and adversarial examples. Although essentially based on the same optimization objective, the author argues that adversarial examples necessitate a misprediction while counterfactual explanations should be different---yet still correct.

While uncertainty quantification in the context of counterfactual explanations remains largely unexplored, we find Delaney \etal \cite{delaneyUncertaintyEstimationOutofDistribution2021} to use  uncertainty quantification methods as a possible measure to increase counterfactual reliability for image classification tasks. In terms of UQ interventions, the authors explore Trust Scores and Monte Carlo dropout, finding Trust Scores to be an effective measure. Schut \etal \cite{schutGeneratingInterpretableCounterfactual2021} propose the direct optimization of an ensemble-based uncertainty measure as a secondary objective for the generation of \textit{realistic} counterfactuals for image classification. In another work, Antorán \etal \cite{antoranGettingCLUEMethod2020} introduce Counterfactual Latent Uncertainty Explanations (CLUE), which is subsequently extended to $\delta$-CLUE \cite{leyDCLUEDiverseSets2021} and GLAM-CLUE \cite{leyDiverseGlobalAmortised2022}. Instead of employing uncertainty quantification to improve counterfactual explainability, CLUE aims to use counterfactual explanations to explain uncertainty estimates in probabilistic models---effectively explaining why certain inputs are more uncertain than others.
% 
%\textbf{Contribution. } Pass we explore the impact 
 
% Est. 3 pages
% ======
% METHOD
% ======
\section{Method}
\label{sec:method}
In this work, we explore the generation of counterfactual samples $x'$ for molecular property prediction tasks, whereby a graph neural network model is trained to regress a continuous property $y$ of a given molecular graph $x$. To gain meaningful insights on the underlying property, we specifically focus on the generation of \textit{truthful} counterfactuals which maximize the prediction difference $|\hat{y} - \hat{y}'|$ between original prediction $\hat{y}$ and counterfactual prediction $\hat{y}'$ while maintaining a minimal ground truth error $|y' - \hat{y}'|$.
%
% task description - graph representation of molecules
%In this study, we specifically focus on molecular property regression as a special subset of graph property prediction tasks. The general objective is to predict a continuous target value $y \in \mathbb{R}$ given an input molecule $x \in \mathcal{X}$.
%
% ===============================
% GRAPH NEURAL NETWORK REGRESSION
% ===============================
%
% Graph neural network definition
\subsection{Graph Neural Network Regressors} We represent each molecule as a generic graph structure $x = (\mathcal{N}, \mathcal{E}, \mathbf{V}^{(0)}, \mathbf{U}^{(0)}) \in \mathcal{X}$ defined by a set of $N$ node indices $\mathcal{N} = \left\{ 1, \dots, N \right\}$ and a list of $E$ edge tuples $\mathcal{E} \subseteq {N} \times \mathcal{N}$ where a tuple $(i, j) \in \mathcal{E}$ indicates an edge between nodes $i$ and $j$. The nodes of this graph structure represent the atoms of the molecule and the edges represent the chemical bonds between the atoms. Furthermore, each graph structure consists of an initial node feature tensor $\mathbf{V}^{(0)} \in \mathbb{R}^{N \times V}$ and an initial edge feature tensor $\mathbf{U}^{(0)} \in \mathbb{R}^{E \times U}$.

In the case of molecular graphs, the node features contain a one-hot encoding of the atom type, the atomic weight, and the charge, whereas the edge features contain a one-hot encoding of the bond type.
% define the GNN learning task on this data structure
\ind For a given dataset of molecules annotated with continuous target values $y \in \mathbb{R}$, the aim is to train a graph neural network regressor 
\begin{equation}
    f_{\theta}: \quad \mathcal{X} \rightarrow \mathbb{R}; \quad (\mathcal{N}, \mathcal{E}, \mathbf{V}^{(0)}, \mathbf{U}^{(0)}) \mapsto \hat{y} 
\end{equation}
with learnable parameters $\theta$ to find an optimal set of parameters 
\begin{equation}
    \theta^{\ast} = \arg\min_{\theta} \sum_{x \in \mathcal{X}_{\text{}}} (y - f_{\theta}(x))^2
\end{equation}
that minimizes the mean-squared error between the predicted value $\hat{y}$ and target $y$ value.
%
% FIGURE: METHOD (AUC / TRUTHFULNESS)
% ===================================
\begin{figure}[t]
    \centering
    \includegraphics[width=1.0\linewidth]{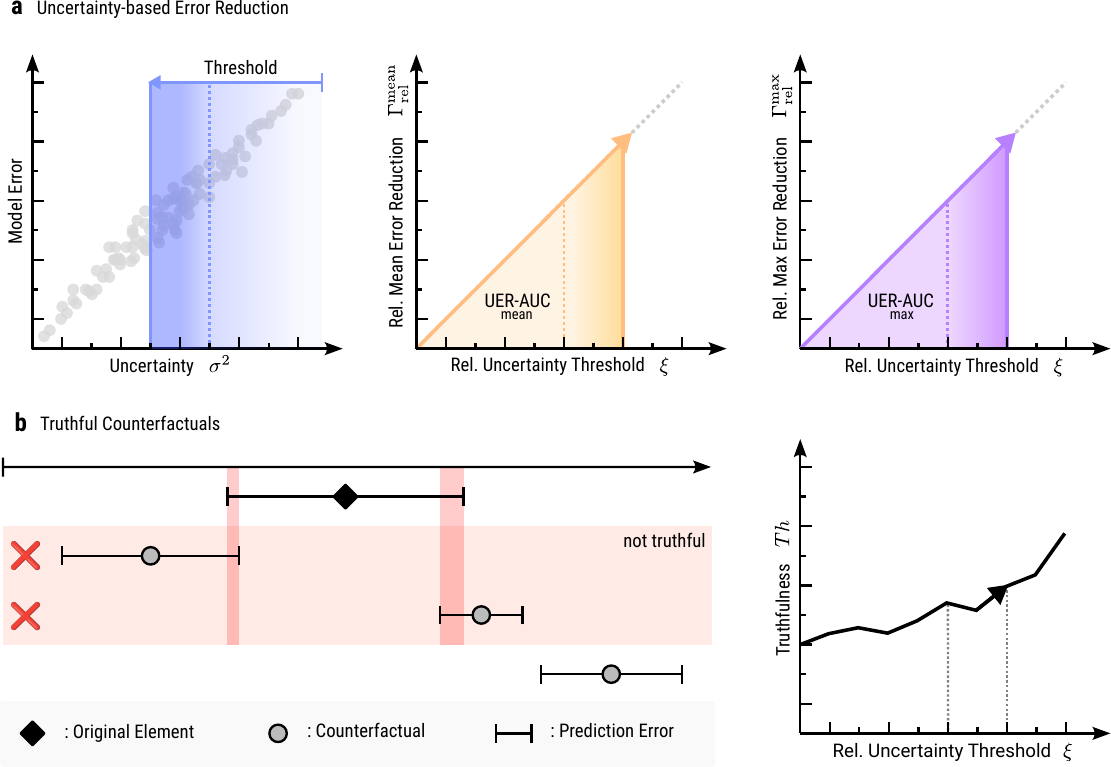}
    \caption{\textbf{\textsf{a}}~Evaluation of the uncertainty-based error reduction over many different thresholds yields characteristic error reduction curves. The Area under the uncertainty error reduction curve (UER-AUC) provides a generic metric for the error reduction potential independent of a specific threshold choice. \textbf{\textsf{b}}~Truthful counterfactuals are defined as those whose prediction error interval does not overlap with that of its corresponding original element. Besides a reduction of the cumulative error, filtering by uncertainty thresholds may also increase the relative fraction of truthful counterfactuals.}
    \label{fig:method}
\end{figure}
%
% ===================================
% MOLECULAR COUNTERFACTUAL GENERATION
% ===================================
%
% general idea behind counterfactuals
\subsection{Molecular Counterfactual Generation }
\label{sec:method-counterfactuals}

Counterfactual explanations map a model's local decision boundary by producing a set of minimally perturbed input instances that induce maximal predictive divergence, thereby revealing which kinds of modifications the model is especially sensitive toward.

% more formal introduction of the objective
Given the combination $(x, \hat{y})$ of an original input element $x$ and its corresponding model prediction $\hat{y}$, a counterfactual sample $(x', \hat{y}')$ consists of input samples $x'$ which are minimally 
\begin{equation}
    \underset{x'}{\min}\; \text{dist}(x', x)
\end{equation}
different from the original input. At the same time, these minimally perturbed input samples should cause a large deviation
\begin{equation}
    \underset{x'}{\max}\; \text{dist}(\hat{y}', \hat{y})
\end{equation}
in the model's prediction.

% How we generate counterfactuals in the regression case
We generate counterfactual samples according to the given constraints by adopting a procedure similar to that presented by Numeroso and Bacciu \cite{numerosoMEGGeneratingMolecular2021}. However, we omit the training of a reinforcement learning agent to induce the local changes of the molecular structure and opt for a complete enumeration of the entire $k$-edit neighborhood instead. Due to the limited number of chemically valid modifications and the relatively small size of molecular graphs, we find it computationally feasible to generate all possible modifications to a given input molecule $x$. As possible modifications, we consider the addition, deletion, and substitution of individual atoms and bonds that satisfy the constraints of atomic valence. Subsequently, the predictive model $f_{\theta}$ is used to obtain the predicted values for all the perturbed graph structures. The structures are then ranked according to the mean absolute prediction difference 
\begin{equation}
    \text{dist}(\hat{y}', \hat{y}) = | \hat{y}' - \hat{y} |
\end{equation}
regarding the original prediction $\hat{y}$. We finally choose the 10 elements with the highest prediction difference to be presented as counterfactual explanations.

At this point, it is worth noting that other possible variations of choosing counterfactual explanations exist. Instead of using the criteria of absolute distance, depending on the use case, it might make sense to select counterfactuals only among those samples with monotonically higher or lower predicted values. 
%
% ===========================
% COUNTERFACTUAL TRUTHFULNESS
% ===========================
%
\subsection{Counterfactual Truthfulness } 
\label{sec:method-truthfulness}

A counterfactual explanation $x'$ has to be a minimal perturbation of the original sample $x$ while causing a large deviation $|\hat{y} - \hat{y}'|$ in the model's prediction. To gain meaningful insight from such counterfactual explanations and to distinguish them from mere adversarial examples \cite{freieslebenIntriguingRelationCounterfactual2022}, we impose the additional restriction of \textit{truthfulness}. We define a truthful counterfactual to additionally maintain a low error $|y' - \hat{y}'|$ regarding its ground truth label $y'$.

For classification problems, we would understand a truthful counterfactual not only to flip the predicted label but to also correctly be associated with that label. For the regression case, there may exist various equally useful definitions of counterfactual truthfulness. In this context, we define a regression counterfactual as truthful if its ground truth error interval does not overlap with the error interval of the original prediction (see Figure~\ref{fig:method}). This definition ensures that there is at least \textit{some} predictive divergence with the predicted directionality.

For a given original sample, its absolute ground truth error 
\begin{equation}
    \epsilon = |y - \hat{y}|
\end{equation}
is calculated as the absolute difference of the true value $y$ and the predicted label $\hat{y}$. The ground truth error
\begin{equation}
    \epsilon' = |y' - \hat{y}'|
\end{equation}
of a counterfactual sample $x'$ can be calculated accordingly. We subsequently define the truthfulness 
\begin{equation}
    \mathrm{tr}(x') = \begin{cases}
        1 & [y' - \epsilon', y' + \epsilon'] \cap [y - \epsilon, y + \epsilon] = \emptyset \\
        0 & \text{otherwise}
    \end{cases}
\end{equation}
of an individual counterfactual as a binary property which is fulfilled if its ground truth error interval does not overlap with the error interval of the original sample.

Beyond the truthfulness of individual counterfactual samples, we are primarily interested in the average truthfulness across a whole set $\mathcal{X}' \subset \mathcal{X}$ of counterfactuals. We, therefore, define the relative truthfulness  
\begin{equation}
    \mathrm{Tr}(\mathcal{X}') = \frac{1}{|\mathcal{X}'|} \sum_{x' \in \mathcal{X}'} \mathrm{tr}(x') \quad \in [0, 1]
\end{equation}
for a set $\mathcal{X}'$ of counterfactuals as the ratio of individual truthful counterfactuals it contains.

At this point, it should be noted that evaluating counterfactual truthfulness proves difficult. Since the generated counterfactual samples generally aren't contained in existing datasets, evaluating the truthfulness would not only require ground truth labels but rather a ground truth oracle. Consequently, truthfulness can only be evaluated for a small selection of tasks for which such an oracle exists.
%
% ===============================
% ERROR REDUCTION BY UNCERTAINTY
% ===============================
%
% General introduction into what kind of error we want to reduce
\subsection{Error Reduction through Uncertainty Thresholding } 
For the given definition of truthfulness, one viable method of improving the relative truthfulness $\mathrm{Tr}(\mathcal{X}')$ is to filter counterfactuals with especially large error intervals. Since it is generally impossible to infer the true label, and by extension the truthfulness, of a given input $x$ in practice, an alternative is to approximate the ground truth error by means of uncertainty quantification (UQ). If the predicted uncertainty proves to be a suitable approximation of the true error, filtering high-uncertainty counterfactuals should have the same effect of improving the relative truthfulness.

This objective can be framed as an overall reduction of the cumulative error 
\begin{equation}
    \Gamma^{g} = g \left( \left\{ | \hat{y}_i - y_i | \,:\, x_i \in \mathcal{X}^{\circ} \right\} \right)
\end{equation}
for a given set $\mathcal{X}^{\circ} \subset \mathcal{X}$ of input elements, where $g(\cdot)$ is some function that accumulates individual error values (e.g. mean, median, max).

In the context of uncertainty quantification, each sample $x$ is additionally assigned a predicted uncertainty value $\sigma^2$. Ideally, a high uncertainty value indicates a potential error in the model prediction, while a low value indicates the prediction to be likely correct. By filtering individual samples with high predicted uncertainties, it should, therefore, be possible to reduce the cumulative error $\Gamma^{g}$ among the remaining elements. For this purpose, we can define the absolute cumulative error 
\begin{equation}
    \Gamma^g(\xi) = g \left( \left\{ | \hat{y}_i - y_i | \, : \, x_i \in \mathcal{X}^{\circ} \; | \; \frac{\sigma^2_i}{\sigma^2_{max}} < \xi  \right\} \right)
\end{equation}
as a function of the relative uncertainty threshold $\xi \in [0, 1]$ used for the filtering.

This definition of cumulative error faces two issues. Firstly, values of the cumulative error will strongly depend on the specific uncertainty threshold $\xi$ that was chosen. 
Secondly, the absolute error scales will be vastly different between different tasks and model performances, therefore not being comparable. Consequently, we propose the \textit{area under the uncertainty error reduction curve (UER-AUC)} as a metric to assess the potential for uncertainty filtering-based error reduction that is comparable across different error scales. To compute the metric, we define the relative cumulative error reduction 
\begin{equation}
    \Delta\Gamma^{g}_{rel}(\xi) = \frac{\Gamma^{g}(1) - \Gamma^{g}(\xi)}{\max_{\xi} \Gamma^{g}(\xi) } \; \in [0, 1]
\end{equation}
which is a value in the range $[0, 1]$, where 0 indicates no error reduction while 1 indicates a 100\% error reduction. We finally define the $\text{UER-AUC}_g$ as the area under the curve of the relative error reduction $\Delta\Gamma^{g}_{rel}(\xi)$ as a function of the relative uncertainty threshold $\xi$. Consequently, the proposed metric is independent of any specific threshold and comparable across different error ranges as both the uncertainty threshold $\xi$ and the relative error reduction $\Delta\Gamma^{g}_{rel}$ are normalized to the range $[0, 1]$.

In terms of accumulation functions $g$, we primarily investigate the mean and the maximum, resulting in the two metrics $\text{UER-AUC}_{\text{mean}}$ and $\text{UER-AUC}_{\text{max}}$. Figure~\ref{fig:method}\textbf{a} illustrates a simple intuition about these metrics: A perfect correlation between uncertainty and model error will result in a UER-AUC of 0.5. Likewise, a UER-AUC of 0 would be the result of a non-existent correlation between uncertainty and error. 

\section{Computational Experiments}

% General introduction
Computational experiments are structured in two major parts: In the first part, we systematically investigate the general error reduction potential of uncertainty estimation methods for different graph neural network architectures, different uncertainty estimation methods, various out-of-distribution settings, and a range of different datasets. In the second part, we consider the use of uncertainty quantification methods in the context of counterfactual explanations and their effect on overall counterfactual truthfulness as previously defined in Section~\ref{sec:method}.
% Introducing the UQ methods
\subsection{Uncertainty Quantification Methods and Metrics}

\subsubsection{Uncertainty Quantification Methods. }
As part of the computational experiments, we compare the following uncertainty quantification methods.
% ~ ensemble
\paragraph{Deep Ensembles (DE).}
We train 3 separate models with bootstrap aggregation, whereby the training data is sampled with replacement. The overall prediction is subsequently obtained as the mean of the individual model outputs, while the standard deviation of the individual predictions is used as an estimate of the uncertainty.
% ~ mve
\paragraph{Mean Variance Estimation (MVE). } The base model architecture is augmented to predict not only the target value $\hat{y}$ but also an uncertainty term $\sigma^2$ by adding additional fully connected layers to the final prediction network \cite{nixEstimatingMeanVariance1994}. The training loss 
\begin{equation}
\label{eq:mve}
\mathcal{L}_{\mathrm{MVE}} = \frac{1}{N} \sum_i^N \frac{\texttt{sg}(\sigma_i^{2\beta})}{2} \cdot \left( \frac{(y_i - \hat{y}_i)^2}{\sigma_i^2} + \log(\sigma_i^2) \right)
\end{equation}
is augmented to optimize both terms at the same time. We specifically integrate the modification proposed by Seitzer \etal \cite{seitzerPitfallsHeteroscedasticUncertainty2022}, which scales the loss by an additional factor of $\sigma^{2\beta}$ but without contributing to the gradient. Furthermore, during training, we follow best practices described by Sluijterman \etal \cite{sluijtermanOptimalTrainingMean2024} by using gradient clipping and including an MSE  warm-up period before switching to the MVE loss. By combining these measures, we substantially improve the performance degradation otherwise reported in the literature.
% ~ ens + mve
\paragraph{Ensemble of mean-variance estimators (DE+MVE). } We combine deep ensembles and mean-variance estimation by constructing an ensemble of 3 independent MVE models, each of which predicts an individual mean and standard deviation, as proposed by Busk \etal \cite{buskCalibratedUncertaintyMolecular2021}. The total uncertainty 
\begin{equation}
\sigma^2 = \frac{1}{2}\left( \sigma^2_{\mathrm{DE}} + \bar{\sigma}^2_{\mathrm{MVE}} \right)
\end{equation}
is calculated as the average of the ensemble uncertainty and the mean MVE uncertainty.
% TABLE: MODEL COMPARISON
% =======================
\begin{table}[t]
    \centering
    \caption{Test set results of 5 independent repetitions of computational experiments on the ClogP dataset regarding different model architectures and uncertainty quantification methods. Normal case numbers are the average result, and lower case gray numbers are the standard deviation. For each combination of model and UQ method, the best results are highlighted in bold, and the second-best results are underlined. }
    {
\small

\setlength{\tabcolsep}{3.5pt}
\begin{tabular}{ llrrrrr }
% -- table header --
\toprule
Model &
UQ Method &
%$\text{MAE} \downarrow$ &
\multicolumn{1}{c}{$R^2 \uparrow$} &
\multicolumn{1}{c}{$\rho \uparrow$} &
\multicolumn{1}{c}{$\underset{mean}{\text{UER-AUC}} \uparrow$} &
\multicolumn{1}{c}{$\underset{max}{\text{UER-AUC}} \uparrow$} & 
\multicolumn{1}{c}{$\text{RLL} \uparrow$}\\

\midrule
% -- table content --
% row 1
--- &
{\smaller Random} &
%$0.13 {\color{gray} \pm \mathsmaller{ 0.05 } }$ &
$1.00 {\color{gray} \pm \mathsmaller{ 0.00 } }$ &
$0.01 {\color{gray} \pm \mathsmaller{ 0.03 } }$ &
$0.01 {\color{gray} \pm \mathsmaller{ 0.04 } }$ &
$0.10 {\color{gray} \pm \mathsmaller{ 0.10 } }$ &
-- 
\\ \arrayrulecolor{gray}\midrule\arrayrulecolor{black}
% -- table content --
% row 1
GCN &
\smaller{DE} &
$1.00 {\color{gray} \pm \mathsmaller{ 0.00 } }$ &
$0.41 {\color{gray} \pm \mathsmaller{ 0.17 } }$ &
$0.21 {\color{gray} \pm \mathsmaller{ 0.06 } }$ &
$0.36 {\color{gray} \pm \mathsmaller{ 0.18 } }$ &
$0.75 {\color{gray} \pm \mathsmaller{ 0.02 } }$ 
\\
% row 3
&
\smaller{MVE} &
$0.99 {\color{gray} \pm \mathsmaller{ 0.01 } }$ &
$0.45 {\color{gray} \pm \mathsmaller{ 0.08 } }$ &
$0.20 {\color{gray} \pm \mathsmaller{ 0.04 } }$ &
$\mathbf{0.63} {\color{gray} \pm \mathsmaller{ 0.18 } }$ &
$\underline{0.77} {\color{gray} \pm \mathsmaller{ 0.03 } }$ 
\\
% row 2
&
\smaller{DE+MVE} &
$1.00 {\color{gray} \pm \mathsmaller{ 0.00 } }$ &
$\mathbf{0.55} {\color{gray} \pm \mathsmaller{ 0.10 } }$ &
$\mathbf{0.25} {\color{gray} \pm \mathsmaller{ 0.02 } }$ &
$\underline{0.58} {\color{gray} \pm \mathsmaller{ 0.20 } }$ &
$\mathbf{0.78} {\color{gray} \pm \mathsmaller{ 0.00 } }$ 
\\
% row 4
&
\smaller{SWAG} &
$1.00 {\color{gray} \pm \mathsmaller{ 0.00 } }$ &
$\underline{0.50} {\color{gray} \pm \mathsmaller{ 0.08 } }$ &
$\underline{0.21} {\color{gray} \pm \mathsmaller{ 0.03 } }$ &
$0.50 {\color{gray} \pm \mathsmaller{ 0.23 } }$ &
$0.56 {\color{gray} \pm \mathsmaller{ 0.11 } }$ 
\\
% row 5
&
{\smaller TS$_{eucl.}$} &
$0.99 {\color{gray} \pm \mathsmaller{ 0.01 } }$ &
$0.15 {\color{gray} \pm \mathsmaller{ 0.17 } }$ &
$0.15 {\color{gray} \pm \mathsmaller{ 0.09 } }$ &
$0.43 {\color{gray} \pm \mathsmaller{ 0.26 } }$ &
$0.69 {\color{gray} \pm \mathsmaller{ 0.10 } }$ 
\\
% row 6
&
{\smaller TS$_{tanim.}$} &
$1.00 {\color{gray} \pm \mathsmaller{ 0.00 } }$ &
$0.15 {\color{gray} \pm \mathsmaller{ 0.05 } }$ &
$0.11 {\color{gray} \pm \mathsmaller{ 0.04 } }$ &
$0.12 {\color{gray} \pm \mathsmaller{ 0.10 } }$ &
$0.39 {\color{gray} \pm \mathsmaller{ 0.33 } }$ 
\\
% sep rule
% -------------------------------------------------
\arrayrulecolor{gray}\midrule\arrayrulecolor{black}
% -- gat --
% row 1
GATv2 &
{\smaller DE} &
$1.00 {\color{gray} \pm \mathsmaller{ 0.00 } }$ &
$\underline{0.51} {\color{gray} \pm \mathsmaller{ 0.11 } }$ &
${0.22} {\color{gray} \pm \mathsmaller{ 0.04 } }$ &
${0.63} {\color{gray} \pm \mathsmaller{ 0.28 } }$ &
$\underline{0.73} {\color{gray} \pm \mathsmaller{ 0.05 } }$ 
\\
% row 2
&
{\smaller MVE} &
% $0.98 {\color{gray} \pm \mathsmaller{ 0.02 } }$ &
% $\mathbf{0.58} {\color{gray} \pm \mathsmaller{ 0.10 } }$ &
% $\mathbf{0.27} {\color{gray} \pm \mathsmaller{ 0.09 } }$ &
% $\underline{0.59} {\color{gray} \pm \mathsmaller{ 0.09 } }$ &
% $\underline{0.68} {\color{gray} \pm \mathsmaller{ 0.06 } }$ 
% new results
$0.98 {\color{gray} \pm \mathsmaller{ 0.03 } }$ &
$0.48 {\color{gray} \pm \mathsmaller{ 0.08 } }$ &
$\underline{0.28} {\color{gray} \pm \mathsmaller{ 0.06 } }$ &
$\underline{0.72} {\color{gray} \pm \mathsmaller{ 0.15 } }$ &
$0.72 {\color{gray} \pm \mathsmaller{ 0.08 } }$ 
\\
&
{\smaller DE+MVE} &
$1.00 {\color{gray} \pm \mathsmaller{ 0.00 } }$ &
$\mathbf{0.64} {\color{gray} \pm \mathsmaller{ 0.15 } }$ &
$\mathbf{0.34} {\color{gray} \pm \mathsmaller{ 0.02 } }$ &
$\mathbf{0.75} {\color{gray} \pm \mathsmaller{ 0.09 } }$ &
$\mathbf{0.82} {\color{gray} \pm \mathsmaller{ 0.03 } }$ 
\\
% row 3
&
{\smaller SWAG} &
$0.99 {\color{gray} \pm \mathsmaller{ 0.00 } }$ &
$0.49 {\color{gray} \pm \mathsmaller{ 0.16 } }$ &
$0.21 {\color{gray} \pm \mathsmaller{ 0.02 } }$ &
$0.61 {\color{gray} \pm \mathsmaller{ 0.21 } }$ &
$-0.06 {\color{gray} \pm \mathsmaller{ 0.47 } }$ 
\\
% row 4
 &
{\smaller TS$_{eucl.}$} &
$1.00 {\color{gray} \pm \mathsmaller{ 0.00 } }$ &
$0.07 {\color{gray} \pm \mathsmaller{ 0.04 } }$ &
$0.17 {\color{gray} \pm \mathsmaller{ 0.07 } }$ &
$0.59 {\color{gray} \pm \mathsmaller{ 0.00 } }$ &
$0.64 {\color{gray} \pm \mathsmaller{ 0.01 } }$ 
\\
% row 5
&
{\smaller TS$_{tanim.}$}&
$1.00 {\color{gray} \pm \mathsmaller{ 0.00 } }$ &
$0.20 {\color{gray} \pm \mathsmaller{ 0.04 } }$ &
$0.13 {\color{gray} \pm \mathsmaller{ 0.03 } }$ &
$0.10 {\color{gray} \pm \mathsmaller{ 0.07 } }$ &
$0.59 {\color{gray} \pm \mathsmaller{ 0.06 } }$ 
\\ 
% sep rule
% -------------------------------------------------
\arrayrulecolor{gray}\midrule\arrayrulecolor{black}
% row 6
GIN &
{\smaller DE} &
$0.99 {\color{gray} \pm \mathsmaller{ 0.01 } }$ &
$\underline{0.62} {\color{gray} \pm \mathsmaller{ 0.17 } }$ &
$\underline{0.27} {\color{gray} \pm \mathsmaller{ 0.06 } }$ &
$\underline{0.70} {\color{gray} \pm \mathsmaller{ 0.11 } }$ &
$\mathbf{0.80} {\color{gray} \pm \mathsmaller{ 0.04 } }$ 
\\
% row 7
&
{\smaller MVE} &
% $0.98 {\color{gray} \pm \mathsmaller{ 0.01 } }$ &
% $\underline{0.58} {\color{gray} \pm \mathsmaller{ 0.14 } }$ &
% $\underline{0.25} {\color{gray} \pm \mathsmaller{ 0.03 } }$ &
% $\underline{0.68} {\color{gray} \pm \mathsmaller{ 0.21 } }$ &
% $\underline{0.71} {\color{gray} \pm \mathsmaller{ 0.03 } }$ 
% new results
$0.99 {\color{gray} \pm \mathsmaller{ 0.01 } }$ &
$0.48 {\color{gray} \pm \mathsmaller{ 0.11 } }$ &
$0.22 {\color{gray} \pm \mathsmaller{ 0.05 } }$ &
$0.56 {\color{gray} \pm \mathsmaller{ 0.22 } }$ &
$0.75 {\color{gray} \pm \mathsmaller{ 0.05 } }$ 
\\
&
{\smaller DE+MVE} &
$1.00 {\color{gray} \pm \mathsmaller{ 0.00 } }$ &
$\mathbf{0.63} {\color{gray} \pm \mathsmaller{ 0.05 } }$ &
$\mathbf{0.29} {\color{gray} \pm \mathsmaller{ 0.03 } }$ &
$\mathbf{0.70} {\color{gray} \pm \mathsmaller{ 0.15 } }$ &
$\underline{0.78} {\color{gray} \pm \mathsmaller{ 0.01 } }$ 
\\
% row 8
 &
{\smaller SWAG} &
$0.98 {\color{gray} \pm \mathsmaller{ 0.02 } }$ &
$0.58 {\color{gray} \pm \mathsmaller{ 0.20 } }$ &
$0.23 {\color{gray} \pm \mathsmaller{ 0.07 } }$ &
$0.58 {\color{gray} \pm \mathsmaller{ 0.08 } }$ &
$0.02 {\color{gray} \pm \mathsmaller{ 0.43 } }$ 
\\
% row 9
 &
{\smaller TS$_{eucl.}$} &
$0.99 {\color{gray} \pm \mathsmaller{ 0.00 } }$ &
$0.15 {\color{gray} \pm \mathsmaller{ 0.12 } }$ &
$0.17 {\color{gray} \pm \mathsmaller{ 0.05 } }$ &
$0.45 {\color{gray} \pm \mathsmaller{ 0.22 } }$ &
$0.64 {\color{gray} \pm \mathsmaller{ 0.03 } }$ 
\\
% row 10
 &
{\smaller TS$_{tanim.}$} &
$0.99 {\color{gray} \pm \mathsmaller{ 0.00 } }$ &
$0.17 {\color{gray} \pm \mathsmaller{ 0.08 } }$ &
$0.13 {\color{gray} \pm \mathsmaller{ 0.05 } }$ &
$0.11 {\color{gray} \pm \mathsmaller{ 0.11 } }$ &
$0.52 {\color{gray} \pm \mathsmaller{ 0.12 } }$ 
\\

\bottomrule
\end{tabular}
}
    \vspace*{2mm}
    \label{tab:models}
\end{table}
%\FloatBarrier
%
% ~ swag
\hspace*{-1.5mm}\paragraph{Stochastic Weight Averaging Gaussian (SWAG). } The training process is augmented to store snapshots of the model weights during the last 25 epochs. This history of model weights is then used to calculate a mean weight vector $\mathbf{\mu}_{\theta}$ and a covariance matrix $\mathbf{\Sigma}_{\theta}$ such that a new set of model weights can approximately be obtained by drawing from a gaussian distribution $\theta \sim \mathcal{N}(\mu_{\theta}, \Sigma_{\theta})$.
During inference, we sample 50 distinct model weights from this distribution and obtain the target value prediction as the mean of the individual predictions and an uncertainty estimate as the standard deviation. 
% ~ trust scores
\paragraph{Trust Scores (TS). } Unlike the previously described UQ methods, trust scores are independent of the predictive model and provide an uncertainty estimate based directly on the training data \cite{delaneyUncertaintyEstimationOutofDistribution2021}. Originally introduced for classification problems, the trust score for a given input element $x$ is calculated as the fraction 
\begin{equation}
T = \frac{\mathrm{dist}(x, x_s)}{\mathrm{dist}(x, x_o)}
\end{equation}
between the distances of the closest training element $x_s$ of the same class and the closest training element $x_o$ of a different class. We adapt this approach for regression tasks by using the distance to the closest element. This definition relies on the existence of a suitable distance metric $\mathrm{dist}(x_i, x_j)$ between two input elements. 
In this study, we examine two distance metrics for comparing input elements. The first is the Tanimoto distance, which is calculated as the Jaccard distance between two Morgan fingerprint representations of two molecules. The second is the Euclidean distance, which is measured between the graph embeddings generated by an intermediate layer of the graph neural network models.
% Calibration
\paragraph{Uncertainty Calibration. } After training, we apply uncertainty calibration to each UQ method to align the predicted uncertainties with the scale of the actual prediction errors. For this purpose, we use a held-out validation set containing 10\% of the data to subsequently fit an isotonic regression model.
% Uncertainty Quantification Metrics
\subsubsection{Uncertainty Quantification Metrics.} We evaluate the aforementioned UQ methods with the following metrics.
\paragraph{Uncertainty-Error Correlation $\rho$.} The Pearson correlation coefficient between the absolute prediction errors $|\hat{y} - y|$ and the predicted uncertainties $\sigma^2$ on the elements of the test set.
\paragraph{Error Reduction Potential UER-AUC.} As described in Section~\ref{sec:method}, the UER-AUC is the area under the curve that maps relative error reduction to relative uncertainty thresholds. For each uncertainty threshold, all elements with higher predicted uncertainty are omitted from the test set. The relative error reduction describes the reduction of the cumulative error of the remaining elements relative to the full set.
\paragraph{Relative Log Likelihood RLL.} Following the work of Kellner and Ceriotti \cite{kellnerUncertaintyQuantificationDirect2024} we use the Relative Log Likelihood 
\begin{equation}
\mathrm{RLL} = \frac{\sum_i \mathrm{NLL}(\hat{y}_i - y_i, \sigma^2_i) - \mathrm{NLL}(\hat{y}_i - y_i, \mathrm{RMSE}) }{\sum_i \mathrm{NLL}(\hat{y}_i - y_i, |\hat{y}_i - y_i|) - NLL(\hat{y}_i - y_i, \mathrm{RMSE}) }
\end{equation}
which standardizes the arbitrary range of the Negative Log Likelihood 
\begin{equation}
\mathrm{NLL}(\Delta y, \sigma^2) = \frac{1}{2} \left( \frac{\Delta y^2}{\sigma^2} + \log 2\pi \sigma^2 \right) 
\end{equation}
into a more interpretable range $(-\infty, 1]$.
%
%
% ================================
% RESULTS PART 1 - ERROR REDUCTION
% ================================
%
\subsection{Experiments on Error Reduction Potential}
%
% ------------------
% INFLUENCE OF MODEL
% ------------------
\subsubsection{Impact of GNN Model and UQ Method on Error Reduction. } In this first experiment, we evaluate the impact of the model architecture and uncertainty quantification method on uncertainty-based error reduction.
% description of the dataset
The experiment is based on the ClogP dataset, which consists of roughly 11k small molecules annotated with values of Crippen's logP \cite{wildmanPredictionPhysicochemicalParameters1999} calculated by RDKit \cite{landrumRDKitOpensourceCheminformatics2010}. This logP value is an algorithmically calculated and deterministic property---making it possible to near-perfectly regress it with machine learning models.

% description of the experiment
In terms of model choice, we compare three standard GNN architectures based on the GCN \cite{kipfSemiSupervisedClassificationGraph2017b}, GATv2 \cite{brodyHowAttentiveAre2022}, and GIN \cite{xu*HowPowerfulAre2018} layer types, respectively. For each repetition of the experiment, we randomly choose 10\% of the dataset as the test set, 10\% as the calibration set and train the model on the remaining. Therefore, the test set can be considered IID w.r.t. to the training distribution.

% discussion of the results
Table~\ref{tab:models} shows the results of the first experiment.
A "Random" baseline, generating random uncertainty values, was included as a control. As expected, this baseline demonstrates negligible error reduction, reflecting the absence of correlation between assigned uncertainty and prediction error. In contrast, the remaining uncertainty quantification methods exhibit varying degrees of error reduction.

Using trust scores with the input-based Tanimoto distance yields substantially worse results than the embedding-based Euclidean distance. Contrary to the encouraging results of Delaney \etal \cite{delaneyUncertaintyEstimationOutofDistribution2021}, we believe trust scores underperform in this particular application due to the challenge of defining suitable distance metrics on graph-structured data \cite{willsMetricsGraphComparison2020}.

Overall, we find deep ensembles, mean-variance estimation, and a combination thereof to work the best in terms of error reduction potential, as well as relative log likelihood. Out of these methods, we observe a slight advantage in mean error reduction for the combined ensemble and mean-variance estimation approach.

Moreover, regarding the different model architectures (GCN, GATv2, and GIN), we observe comparable results, both in terms of predictive performance ($R^2\geq0.99$) and in terms of uncertainty quantification methods. Based on these initial observations, model architecture appears to have a limited effect on the relative performance of the uncertainty quantification methods. Consequently, subsequent experiments were conducted using the GATv2 architecture, which exhibited the highest mean error reduction potential in this experiment.

% In terms of UQ methods, Ensembling and MVE without calibration show comparable results regarding the potential for mean error reduction ($\aucmean \approx 0.2$) while MVE shows slightly higher max error reduction. Another noteworthy effect is a slight decrease of prediction performance for the MVE method---likely caused by occasional destabilization of the model training due to the additional MVE loss term \cite{sluijtermanOptimalTrainingMean2024}.\\
% While the additional uncertainty calibration seemingly did not affect the Ensembling results, it provided a significant improvement for the MVE method in all aspects achieving up to 50\% increase in average error reduction potential ($\aucmean \approx 0.3$).\\
% Regarding the choice of model architecture, we observe similar results for the exemplary GAT and GIN-based models. Both models show similar results for the uncalibrated MVE and Ensembling method, as well as a substantial improvement in the case of the calibrated MVE method.\\
% Overall, the results of this experiment indicate that the choice of network architecture seemingly has little effect on error reduction and that uncertainty calibration can have a positive effect in some cases. For subsequent experiments, we therefore only use the GATv2 model and always apply uncertainty calibration.
% =======================
% FIGURE: EXAMPLE RESULTS
% =======================
\begin{figure}[t]
    \centering
    \includegraphics[width=\linewidth]{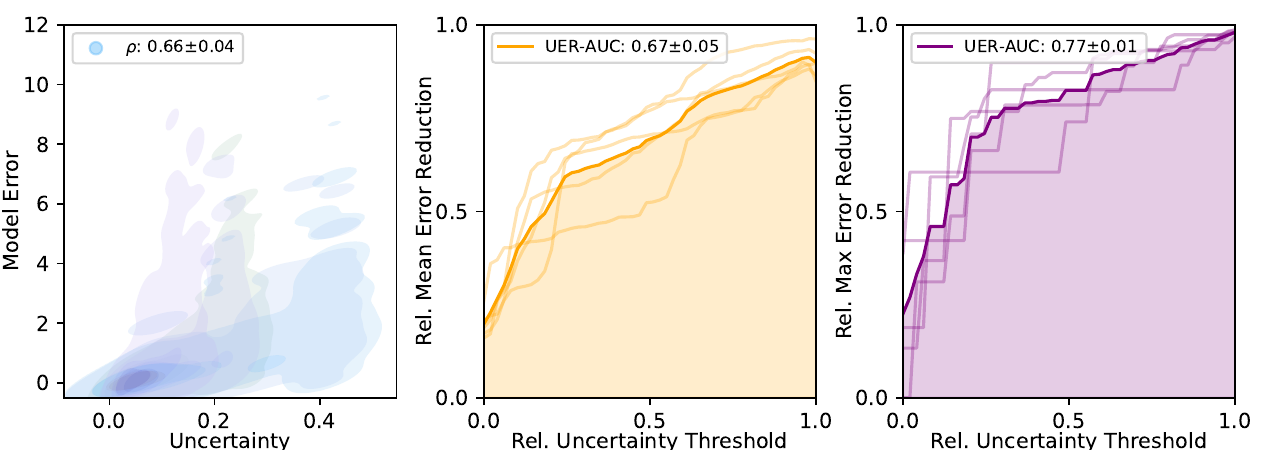}
    \caption{Results for 5 independent repetitions of a GATv2 trained on the ClogP dataset and uncertainties estimated with a combination of ensembles and mean variance estimation in the OOD-Value scenario. Panels from left to right illustrate the correlation between the predicted uncertainty \& model error, the mean error reduction potential, and the max error reduction potential through filtering by uncertainty thresholds. Faint lines represent the results of individual runs; bold lines represent the overall average.}
    \label{fig:example}
\end{figure}
%\FloatBarrier
%
% ---------------------------
% OUT OF DISTRIBUTION EFFECTS
% ---------------------------
\subsubsection{Out-of-distribution Effect on Error Reduction. } The previous experiment examined the error reduction potential on a randomly sampled IID test set of the ClogP dataset. However, a critical aspect of counterfactual analysis involves identifying input perturbations that yield out-of-distribution (OOD) samples. To address this, we established two OOD scenarios for the ClogP dataset. The first, designated \textit{OOD-Struct}, employs a scaffold split, where the test set comprises molecules with structural scaffolds absent from the training set. The second, \textit{OOD-Value}, involves a split where the test set contains approximately the 10\% most extreme target values, not represented in the training set. Due to the results of the previous experiments, for each scenario, we restrict experiments to use the GATv2 model architecture and compare uncertainty estimation based on ensembles, mean variance estimation, and the combination thereof.

% ==========================
% TABLE: OUT OF DISTRIBUTION
% ==========================
\begin{table}[t]
    \caption{Test set results of 5 independent repetitions of computational experiments on the ClogP dataset regarding different out-of-distribution scenarios and uncertainty quantification methods. The best result for each scenario is highlighted in bold, and the second-best result is underlined. Results were obtained based on a GATv2 model architecture.}
    \label{tab:ood}
    {
\small
\centering
\setlength{\tabcolsep}{2.9pt}
\begin{tabular}{ llrrrrr }
% -- table header --
\toprule
Scenario &
UQ Method &
\multicolumn{1}{c}{$R^2 \uparrow$} &
\multicolumn{1}{c}{$\rho \uparrow$} &
\multicolumn{1}{c}{$\underset{mean}{\text{UER-AUC}} \uparrow$} &
\multicolumn{1}{c}{$\underset{max}{\text{UER-AUC}} \uparrow$} & 
\multicolumn{1}{c}{$\text{RLL} \uparrow$}
\\

\midrule

% OMITTED: IID VALUES
% % -- table content --
% % row 1
% IID &
% {\smaller DE} &
% $\mathbf{1.00} {\color{gray} \pm \mathsmaller{ 0.00 } }$ &
% $\underline{0.57} {\color{gray} \pm \mathsmaller{ 0.12 } }$ &
% $\underline{0.25} {\color{gray} \pm \mathsmaller{ 0.05 } }$ &
% $\mathbf{0.69} {\color{gray} \pm \mathsmaller{ 0.21 } }$ &
% $\mathbf{0.76} {\color{gray} \pm \mathsmaller{ 0.04 } }$ 
% \\
% % row 3
%  &
% {\smaller MVE} &
% $0.98 {\color{gray} \pm \mathsmaller{ 0.01 } }$ &
% $0.40 {\color{gray} \pm \mathsmaller{ 0.10 } }$ &
% $0.19 {\color{gray} \pm \mathsmaller{ 0.09 } }$ &
% $0.61 {\color{gray} \pm \mathsmaller{ 0.21 } }$ &
% $0.67 {\color{gray} \pm \mathsmaller{ 0.04 } }$ 
% \\
% % row 2
% &
% {\smaller DE+MVE} &
% $\underline{0.98} {\color{gray} \pm \mathsmaller{ 0.01 } }$ &
% $\mathbf{0.64} {\color{gray} \pm \mathsmaller{ 0.19 } }$ &
% $\mathbf{0.30} {\color{gray} \pm \mathsmaller{ 0.09 } }$ &
% $\underline{0.66} {\color{gray} \pm \mathsmaller{ 0.11 } }$ &
% $\underline{0.76} {\color{gray} \pm \mathsmaller{ 0.06 } }$ 
% \\
% % sep rule
% % -------------------------------------------------
% \arrayrulecolor{gray}\midrule\arrayrulecolor{black}

% row 4
OOD &
{\smaller DE} &
$1.00 {\color{gray} \pm \mathsmaller{ 0.00 } }$ &
$\underline{0.45} {\color{gray} \pm \mathsmaller{ 0.05 } }$ &
$\mathbf{0.23} {\color{gray} \pm \mathsmaller{ 0.04 } }$ &
$\mathbf{0.66} {\color{gray} \pm \mathsmaller{ 0.07 } }$ &
$\underline{0.34} {\color{gray} \pm \mathsmaller{ 0.06 } }$ 
\\
% row 6
struct&
{\smaller MVE} &
$0.99 {\color{gray} \pm \mathsmaller{ 0.00 } }$ &
$0.32 {\color{gray} \pm \mathsmaller{ 0.15 } }$ &
$0.14 {\color{gray} \pm \mathsmaller{ 0.06 } }$ &
$0.20 {\color{gray} \pm \mathsmaller{ 0.11 } }$ &
$0.41 {\color{gray} \pm \mathsmaller{ 0.03 } }$ 
\\
% row 5
&
\smaller DE+MVE &
$1.00 {\color{gray} \pm \mathsmaller{ 0.00 } }$ &
$\mathbf{0.46} {\color{gray} \pm \mathsmaller{ 0.05 } }$ &
$\underline{0.21} {\color{gray} \pm \mathsmaller{ 0.04 } }$ &
$\underline{0.42} {\color{gray} \pm \mathsmaller{ 0.17 } }$ &
$\mathbf{0.55} {\color{gray} \pm \mathsmaller{ 0.00 } }$ 
\\
% sep rule
% -------------------------------------------------
\arrayrulecolor{gray}\midrule\arrayrulecolor{black}
% row 7
OOD &
{\smaller DE} &
$0.97 {\color{gray} \pm \mathsmaller{ 0.01 } }$ &
$\underline{0.62} {\color{gray} \pm \mathsmaller{ 0.07 } }$ &
$\mathbf{0.71} {\color{gray} \pm \mathsmaller{ 0.10 } }$ &
$\mathbf{0.82} {\color{gray} \pm \mathsmaller{ 0.07 } }$ &
$\underline{-3.79} {\color{gray} \pm \mathsmaller{ 2.92 } }$ 
\\
% row 9
value &
{\smaller MVE} &
$0.99 {\color{gray} \pm \mathsmaller{ 0.00 } }$ &
$0.50 {\color{gray} \pm \mathsmaller{ 0.08 } }$ &
$0.51 {\color{gray} \pm \mathsmaller{ 0.11 } }$ &
$0.36 {\color{gray} \pm \mathsmaller{ 0.27 } }$ &
$-8.04 {\color{gray} \pm \mathsmaller{ 9.72 } }$ 
\\
% row 8
 &
{\smaller DE+MVE} &
$0.98 {\color{gray} \pm \mathsmaller{ 0.00 } }$ &
$\mathbf{0.66} {\color{gray} \pm \mathsmaller{ 0.04 } }$ &
$\underline{0.67} {\color{gray} \pm \mathsmaller{ 0.05 } }$ &
$\underline{0.77} {\color{gray} \pm \mathsmaller{ 0.01 } }$ &
$\mathbf{-1.49} {\color{gray} \pm \mathsmaller{ 1.09 } }$ 
\\

\bottomrule
\end{tabular}
}
    \vspace*{2mm}
\end{table}
%
% Discussing the results
Table~\ref{tab:ood} reports the results of the second experiment. For the OOD-Struct scenario, we observe slightly worse results than for the IID case. All three methods show lower correlation, error reduction potential, and relative log likelihood. 

Conversely, the OOD-Value scenario exhibited substantial performance gains relative to the IID case. Mean and max error reduction potential cross decisively  
exceed the $\text{UER-AUC}\geq0.5$ threshold. Only the negative RLL values indicate poorly calibrated uncertainty estimates with respect to the actual prediction error. This is to be expected since the calibration set was sampled IID while the test set contains previously unseen target values---likely resulting in vastly different error scales.

When comparing the different UQ methods, the ensembles by themselves and the combination of ensembles and MVE seem to perform equally well. For both scenarios, OOD-struct and OOD-value, the ensembles seem to offer higher error reduction potential, while the combination seems to offer better calibrated uncertainty estimates, as indicated by the higher RLL values.

In summary, uncertainty-based filtering demonstrates a moderate error reduction effect on in-distribution data and structural outliers. Notably, the error reduction potential increases substantially under a distributional shift of the target values (see Figure~\ref{fig:example}). These results provide a foundation for filtering counterfactuals, where perturbations can be expected to create outliers with respect to both structure and target value.

\begin{table}[t]
    \centering
    \caption{Test set results of 5 independent repetitions of computational experiments to evaluate uncertainty-based error reduction on various molecular property prediction datasets. The first row represents the previously introduced deterministic CLogP graph regression task, and the following rows represent various real-world molecular property regression datasets. Results are obtained by a GATv2 graph neural network and uncertainties are estimated by a method combining deep ensembles and mean-variance estimation.}
    \label{tab:real}
    {
\small
\setlength{\tabcolsep}{2.7pt}
\begin{tabular}{ llrrrrr }
% -- table header --
\toprule
Dataset &
Property &
\multicolumn{1}{c}{$R^2 \uparrow$} &
\multicolumn{1}{c}{$\rho \uparrow$} &
\multicolumn{1}{c}{$\underset{mean}{\text{UER-AUC}} \uparrow$} &
\multicolumn{1}{c}{$\underset{max}{\text{UER-AUC}} \uparrow$} &
\multicolumn{1}{c}{$\text{RLL} \uparrow$} \\

\midrule

% -- table content --
% row 1
ClogP &
logP &
$1.00 {\color{gray} \pm \mathsmaller{ 0.00 } }$ &
$0.58 {\color{gray} \pm \mathsmaller{ 0.17 } }$ &
$0.27 {\color{gray} \pm \mathsmaller{ 0.05 } }$ &
$0.66 {\color{gray} \pm \mathsmaller{ 0.22 } }$ &
$0.76 {\color{gray} \pm \mathsmaller{ 0.02 } }$  
\\
% sep rule
% -------------------------------------------------
\arrayrulecolor{gray}\midrule\arrayrulecolor{black}
% row 2
AqSolDB\cite{sorkunAqSolDBCuratedReference2019} &
logS &
$0.88 {\color{gray} \pm \mathsmaller{ 0.02 } }$ &
$0.35 {\color{gray} \pm \mathsmaller{ 0.05 } }$ &
$0.24 {\color{gray} \pm \mathsmaller{ 0.02 } }$ &
$0.26 {\color{gray} \pm \mathsmaller{ 0.17 } }$ &
$0.45 {\color{gray} \pm \mathsmaller{ 0.03 } }$ 
\\
% row 3
Lipop\cite{wuMoleculeNetBenchmarkMolecular2018} &
logD &
$0.74 {\color{gray} \pm \mathsmaller{ 0.03 } }$ &
$0.15 {\color{gray} \pm \mathsmaller{ 0.06 } }$ &
$0.10 {\color{gray} \pm \mathsmaller{ 0.02 } }$ &
$0.22 {\color{gray} \pm \mathsmaller{ 0.12 } }$ &
$0.32 {\color{gray} \pm \mathsmaller{ 0.02 } }$ 
\\
% row 4
COMPAS\cite{wahabCOMPASProjectComputational2022} &
rel. Ener. &
$0.90 {\color{gray} \pm \mathsmaller{ 0.05 } }$ &
$0.65 {\color{gray} \pm \mathsmaller{ 0.04 } }$ &
$0.37 {\color{gray} \pm \mathsmaller{ 0.03 } }$ &
$0.45 {\color{gray} \pm \mathsmaller{ 0.11 } }$ &
$0.66 {\color{gray} \pm \mathsmaller{ 0.03 } }$
\\
% row 5
 &
GAP &
$0.97 {\color{gray} \pm \mathsmaller{ 0.01 } }$ &
$0.44 {\color{gray} \pm \mathsmaller{ 0.05 } }$ &
$0.27 {\color{gray} \pm \mathsmaller{ 0.05 } }$ &
$0.59 {\color{gray} \pm \mathsmaller{ 0.20 } }$ &
$0.71 {\color{gray} \pm \mathsmaller{ 0.01 } }$ 
\\
% row 6
QM9\cite{ramakrishnanQuantumChemistryStructures2014} &
Dip. Mom. &
$0.78 {\color{gray} \pm \mathsmaller{ 0.00 } }$ &
$0.57 {\color{gray} \pm \mathsmaller{ 0.01 } }$ &
$0.45 {\color{gray} \pm \mathsmaller{ 0.01 } }$ &
$0.76 {\color{gray} \pm \mathsmaller{ 0.03 } }$ &
$0.53 {\color{gray} \pm \mathsmaller{ 0.00 } }$ 
\\
% row 7
 &
HOMO &
$0.93 {\color{gray} \pm \mathsmaller{ 0.00 } }$ &
$0.54 {\color{gray} \pm \mathsmaller{ 0.02 } }$ &
$0.23 {\color{gray} \pm \mathsmaller{ 0.01 } }$ &
$0.61 {\color{gray} \pm \mathsmaller{ 0.10 } }$ &
$0.63 {\color{gray} \pm \mathsmaller{ 0.01 } }$  
\\
% row 8
 &
LUMO &
$0.99 {\color{gray} \pm \mathsmaller{ 0.00 } }$ &
$0.48 {\color{gray} \pm \mathsmaller{ 0.02 } }$ &
$0.23 {\color{gray} \pm \mathsmaller{ 0.01 } }$ &
$0.67 {\color{gray} \pm \mathsmaller{ 0.02 } }$ &
$0.73 {\color{gray} \pm \mathsmaller{ 0.00 } }$ 
\\
% row 9
 &
GAP &
$0.97 {\color{gray} \pm \mathsmaller{ 0.00 } }$ &
$0.52 {\color{gray} \pm \mathsmaller{ 0.02 } }$ &
$0.25 {\color{gray} \pm \mathsmaller{ 0.01 } }$ &
$0.76 {\color{gray} \pm \mathsmaller{ 0.04 } }$ &
$0.68 {\color{gray} \pm \mathsmaller{ 0.00 } }$
\\

\bottomrule
\end{tabular}
}
    \vspace*{2mm}
\end{table}
%\FloatBarrier
%====================================
% FIGURE: SOME COUNTERFACTUALS
% ===================================
\begin{figure}[t]
    \centering
    \includegraphics[width=1.0\linewidth]{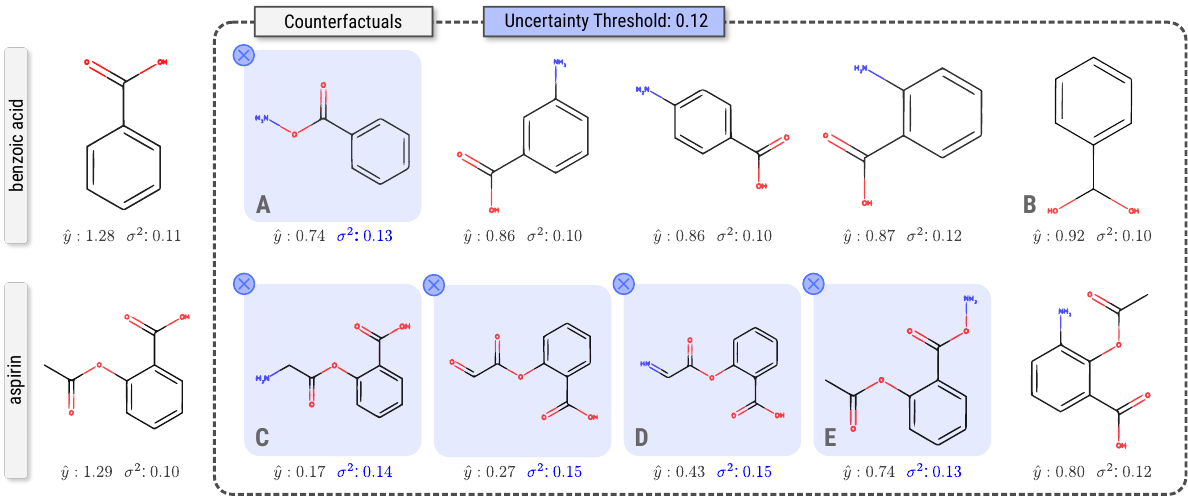}
    \caption{Qualitative results of uncertainty-based counterfactual filtering for two example molecules. Predictions are made by a GATv2 graph neural network and uncertainties are estimated by a combination of ensembling and mean-variance estimation. The uncertainty threshold $\xi_{20}$ was chosen on the test set such that the 20\% lowest uncertainty elements remain.}
    \label{fig:qual}
\end{figure}
% -------------------
% REAL WORLD DATASETS
% -------------------
% Introduction of the experiment
\subsubsection{Error Reduction on Real-World Datasets. } Previous experiments were based on the ClogP dataset, which is a deterministically computable property and, therefore, relatively easy to regress. To assess the generalizability of these findings to more complex scenarios, computational experiments were conducted on multiple properties derived from the AqSolDB \cite{sorkunAqSolDBCuratedReference2019}, Lipop \cite{wuMoleculeNetBenchmarkMolecular2018}, COMPAS \cite{wahabCOMPASProjectComputational2022}, and QM9 \cite{ramakrishnanQuantumChemistryStructures2014} datasets.  Based on the results of previous experiments, we use the GATv2 model to predict each property and a combination of ensembles and mean-variance estimation for the uncertainty quantification.

% Discussion of the results
Table~\ref{tab:real} presents the results for the real-world property regression datasets. Despite varying levels of predictivity ($R^2 \in [0.74, 0.99]$) for the different datasets, some degree of error reduction can be reported for each one ($\aucmean \in [0.10, 0.45]$). Notably, the highest error reduction is found for the prediction of the Dipole Moment in the QM9 dataset with a mean error reduction of $\aucmean=0.45$ and a max error reduction of $\aucmax=0.78$. In contrast, the lowest error reduction can be observed for the prediction of the Lipophilicity with a mean error reduction of only $\aucmean=0.10$.

The extent of error reduction potential does not appear to correlate strongly with the predictive performance of the model, as both the highest and lowest error reductions were associated with models demonstrating lower predictivity ($R^2 \approx 0.7$). In addition, even models with high predictivity, such as the prediction of the LUMO value ($R^2 = 0.99$), show moderate amounts of error reduction potential ($\aucmean=0.23$, $\aucmax=0.67$). We hypothesize that the error reduction may be connected to the complexity of the underlying data distribution and the presence of labeling noise. The Lipophilicity dataset, for example, consists of inherently noisy experimental measurements while values for the dipole moment in the QM9 dataset were obtained by more precise DFT simulations.

% Conclusion of the experiment
Overall, the results of this experiment indicate that uncertainty threshold-based filtering can be used as an effective tool to decrease the overall prediction error even for complex properties, which may have been obtained through noisy measurements.

\subsection{Experiments on Counterfactual Truthfulness}
% small introduction to the second part
In the second part of the computational experiments, we investigate the potential of uncertainty-based filtering to improve the overall truthfulness of counterfactual explanations.

% -------------------------
% TRUTHFULNESS QUANTITATIVE
% -------------------------
\subsubsection{Improving Counterfactual Truthfulness. } 

% description of the experiment
For this experiment, we use the CLogP dataset, as the underlying property is deterministically calculable for any valid molecular graph. This availability of a ground truth oracle is necessary for the computation of the relative truthfulness as defined in Section~\ref{sec:method-truthfulness}. As before, we use the GATv2 model architecture and investigate the effectiveness of ensembles, mean-variance estimation, and the combination thereof. We split the dataset into a test set (10\%), a calibration set (20\%), and a train set (70\%). All models are fitted with the train set, and uncertainty estimates are subsequently calibrated on the validation set. On the test set, we determine a single uncertainty threshold $\xi_{20}$ such that exactly the 20\% elements with the lowest predicted uncertainties remain.

As described in Section~\ref{sec:method-counterfactuals}, we generate counterfactual samples by ranking all graphs in a 1-edit neighborhood according to the prediction divergence and choosing the top 10 candidates. This set of counterfactuals is then filtered using the threshold $\xi_{20}$ and examined regarding its relative truthfulness.

% description of the results
The results of this experiment are reported in Table~\ref{tab:cf}. A "Random" baseline was included as a control. As expected, this control's randomly generated uncertainty values result neither in test set error reduction nor an increase of counterfactual truthfulness. All other uncertainty quantification (UQ) methods demonstrated a moderate potential for error reduction on both the test set and the set of counterfactuals ($\text{UER-AUC} \geq 0.2$). Furthermore, all UQ methods exhibited some capability to increase relative truthfulness when filtering with the uncertainty threshold $\xi_{20}$. However, it has to be noted that the initial truthfulness in the unfiltered set of counterfactuals is rather high (up to 95\%), leaving little room for further improvement. It is important to note, however, that the initial truthfulness in the unfiltered set of counterfactuals was relatively high (up to 95\%), limiting the scope for further improvement. Notably, the mean variance estimation model displayed a substantially lower initial truthfulness (0.77), most likely due to its slightly lower predictivity.

\begin{figure}[t]
    \centering
    \includegraphics[width=1.0\linewidth]{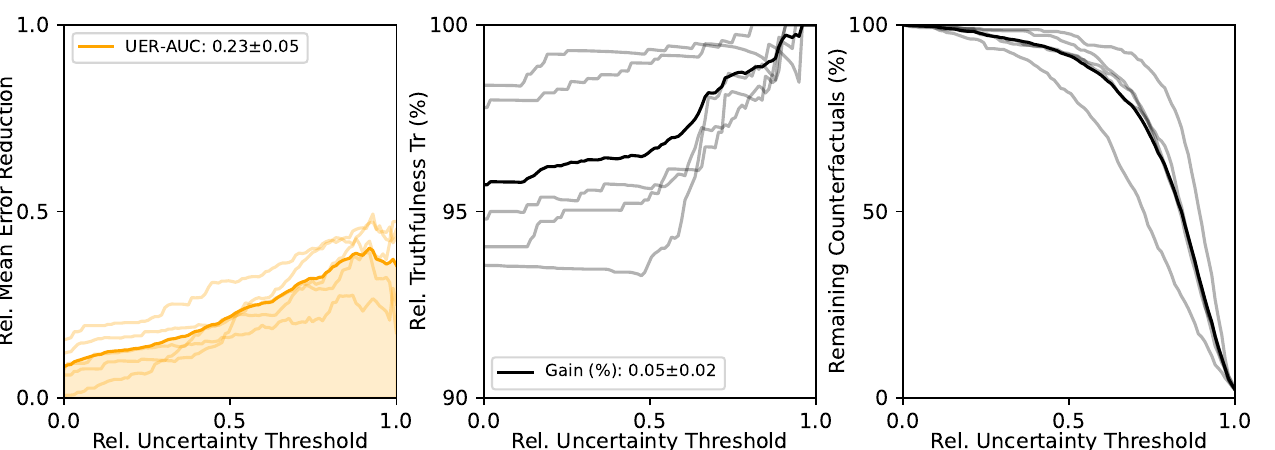}
    \caption{Results for 5 independent repetitions of computational experiment on the ClogP dataset to evaluate counterfactual truthfulness. Individual results are plotted transparently in the background, and average curves are indicated with bold lines. All plots are based on the set of counterfactuals and show from left to right: The relative mean error reduction, the Truthfulness, and the percentage of remaining counterfactuals for different uncertainty thresholds. Results are obtained by a GATv2 graph neural network, and uncertainties are estimated by an ensemble of mean variance estimators.}
    \label{fig:cf}
\end{figure}

% description of the figure
In addition to the results for the fixed uncertainty threshold $\xi_{20}$, Figure~\ref{fig:cf} visualizes the progression of mean error reduction and truthfulness results across a range of possible uncertainty thresholds. The plots show that for increasingly strict uncertainty thresholds, the relative counterfactual truthfulness also increases near-monotonically, reaching 100\% with a small subset of 5\% remaining counterfactuals.

% discussion
In summary, we find that all UQ methods exhibit some capacity to improve the relative counterfactual truthfulness through uncertainty-based filtering. However, the results may be influenced by the already elevated values observed for the unfiltered set. Future work should explore more complex property prediction tasks with lower predictive performance and, consequently, lower starting points of counterfactual truthfulness.

\begin{table}[t]
    \centering
    \caption{Results of 5 independent repetitions of computational experiments on the ClogP dataset to evaluate counterfactual truthfulness using a fixed uncertainty threshold $\xi_{20}$ determined on the test set. Results are obtained using a GATv2 graph neural network, and uncertainties are estimated by the combination of Deep Ensembles and MVE. $^\dagger$\textit{Tr. Init.} represents the initial percentage of truthful counterfactuals in the unfiltered set of all counterfactuals. $^\ddagger$\textit{Tr. Gain.} shows the increase in the relative percentage of truthful counterfactuals after filtering according to the uncertainty threshold $\xi_{20}$.}
    \label{tab:cf}
    \small
\setlength{\tabcolsep}{2.8pt}
\begin{tabular}{ lrrrrrr }
% -- table header --
\toprule
\multicolumn{1}{c}{Method} & 
\multicolumn{2}{c}{Test Set} &
\multicolumn{4}{c}{Counterfactuals} \\
\cmidrule(l){2-3} \cmidrule(l){4-7}
 &
\multicolumn{1}{c}{$R^2 \uparrow$} &
\multicolumn{1}{c}{$\underset{\text{mean}}{\text{UER-AUC}} \uparrow$} &
\multicolumn{1}{c}{$\rho \uparrow$} &
\multicolumn{1}{c}{$\underset{\text{mean}}{\text{UER-AUC}} \uparrow$} &
\multicolumn{1}{c}{$\underset{(\%)}{\text{Tr. Init.}^\dagger} \uparrow$} &
\multicolumn{1}{c}{$\underset{(\%)}{\text{Tr. Gain}^\ddagger} \uparrow$} \\

\midrule
% -- table content --
% row 2
{\smaller Random} &
$1.00 {\color{gray} \pm \mathsmaller{ 0.01 } }$ &
$-0.00 {\color{gray} \pm \mathsmaller{ 0.06 } }$ &
$-0.01 {\color{gray} \pm \mathsmaller{ 0.03 } }$ &
$-0.02 {\color{gray} \pm \mathsmaller{ 0.03 } }$ &
$0.95 {\color{gray} \pm \mathsmaller{ 0.03 } }$ &
$-0.00 {\color{gray} \pm \mathsmaller{ 0.05 } }$
\\
\arrayrulecolor{gray}\midrule\arrayrulecolor{black}

% row 1
{\smaller MVE} &
$0.98 {\color{gray} \pm \mathsmaller{ 0.02 } }$ &
$0.26 {\color{gray} \pm \mathsmaller{ 0.05 } }$ &
$0.53 {\color{gray} \pm \mathsmaller{ 0.15 } }$ &
$0.23 {\color{gray} \pm \mathsmaller{ 0.05 } }$ &
$0.77 {\color{gray} \pm \mathsmaller{ 0.17 } }$ &
$0.09 {\color{gray} \pm \mathsmaller{ 0.05 } }$ 
\\
% row 2
{\smaller DE} &
$1.00 {\color{gray} \pm \mathsmaller{ 0.00 } }$ &
$0.27 {\color{gray} \pm \mathsmaller{ 0.04 } }$ &
$0.45 {\color{gray} \pm \mathsmaller{ 0.12 } }$ &
$0.20 {\color{gray} \pm \mathsmaller{ 0.05 } }$ &
$0.94 {\color{gray} \pm \mathsmaller{ 0.03 } }$ &
$0.04 {\color{gray} \pm \mathsmaller{ 0.05 } }$ 
\\
% row 3
{\smaller DE+MVE} &
$1.00 {\color{gray} \pm \mathsmaller{ 0.00 } }$ &
$0.28 {\color{gray} \pm \mathsmaller{ 0.03 } }$ &
$0.44 {\color{gray} \pm \mathsmaller{ 0.16 } }$ &
$0.23 {\color{gray} \pm \mathsmaller{ 0.05 } }$ &
$0.95 {\color{gray} \pm \mathsmaller{ 0.02 } }$ &
$0.05 {\color{gray} \pm \mathsmaller{ 0.02 } }$ 
\\

% ~ pre MVE fix
% % row 1
% {\smaller MVE} &
% $0.98 {\color{gray} \pm \mathsmaller{ 0.01 } }$ &
% $0.16 {\color{gray} \pm \mathsmaller{ 0.06 } }$ &
% $0.29 {\color{gray} \pm \mathsmaller{ 0.19 } }$ &
% $0.10 {\color{gray} \pm \mathsmaller{ 0.09 } }$ &
% $0.76 {\color{gray} \pm \mathsmaller{ 0.09 } }$ &
% $0.08 {\color{gray} \pm \mathsmaller{ 0.10 } }$ 
% \\
% % row 2
% {\smaller DE} &
% $1.00 {\color{gray} \pm \mathsmaller{ 0.00 } }$ &
% $0.29 {\color{gray} \pm \mathsmaller{ 0.03 } }$ &
% $0.35 {\color{gray} \pm \mathsmaller{ 0.18 } }$ &
% $0.19 {\color{gray} \pm \mathsmaller{ 0.06 } }$ &
% $0.97 {\color{gray} \pm \mathsmaller{ 0.01 } }$ &
% $0.03 {\color{gray} \pm \mathsmaller{ 0.02 } }$ 
% \\
% % row 3
% {\smaller DE+MVE} &
% $0.97 {\color{gray} \pm \mathsmaller{ 0.03 } }$ &
% $0.33 {\color{gray} \pm \mathsmaller{ 0.02 } }$ &
% $0.69 {\color{gray} \pm \mathsmaller{ 0.15 } }$ &
% $0.27 {\color{gray} \pm \mathsmaller{ 0.05 } }$ &
% $0.73 {\color{gray} \pm \mathsmaller{ 0.21 } }$ &
% $0.18 {\color{gray} \pm \mathsmaller{ 0.11 } }$ 
% \\

\bottomrule
\end{tabular}
    \vspace*{2mm}
\end{table}
%%\FloatBarrier
% summary / interpretation
% In total, the computational experiment suggests that filtering elements based on uncertainty thresholds is a suitable method to improve counterfactual truthfulness. On average, we observe an overall positive trend in Truthfulness for increasingly strict thresholds. However, the results indicate that a substantial amount of elements has to be removed to achieve a significant improvement.\\[2mm]
%
% ===================================
% FIGURE: COUNTERFACTUAL TRUTHFULNESS
% ===================================
%
% ------------------------
% TRUTHFULNESS QUALITATIVE
% ------------------------
\subsubsection{Qualitative Filtering Results. } 

Besides a quantitative evaluation of counterfactual truthfulness, some qualitative results of uncertainty-based filtering are illustrated in Figure~\ref{fig:qual} for two example molecules. Uncertainty estimates are obtained by an ensemble of mean variance estimators. As before, the uncertainty threshold $\xi_{20}$ is chosen such that only 20\% of the test set elements with the lowest uncertainty values remain.

% benzoic acid
For the first molecule, benzoic acid, only the highest-ranked counterfactual candidate \textbf{\textsf{\small A}} is filtered due to exceeding the uncertainty threshold. This exclusion intuitively makes sense since the added bond between oxygen and nitrogen is an uncommon configuration that is not represented in the underlying dataset and can be considered an out-of-distribution input. In contrast to this expected behavior, the counterfactual candidate \textbf{\textsf{\small B}} is not filtered but represents an equally uncommon configuration with one carbon being connected to two single-bonded oxygen at the same time. Notably, the model also predicts a significantly incorrect value for this counterfactual candidate.

% aspirin
For the second molecule, aspirin, the four highest-ranked counterfactuals are filtered based on their predicted uncertainty. The exclusion of the counterfactual candidate \textbf{\textsf{\small E}} also intuitively makes sense since it includes the same uncommon bond between nitrogen and oxygen. Excluded counterfactual candidate \textbf{\textsf{\small D}} also contains a rather uncommon substructure but, more importantly, is predicted highly inaccurately by the model. In contrast to these cases, the highest ranked counterfactual candidate \textbf{\textsf{\small C}} is excluded even though the model's prediction is highly accurate, serving as an example of overly conservative filtering.

% discussion
% some exclusions are correct, but there are failure modes of this method where some mistakes where the uncertainty filtering is too conservative in some cases and not careful enough in other cases.
Overall, the qualitative examples illustrate that the uncertainty-based filtering can be effective in identifying and removing out-of-distribution input samples and generally inaccurate predictions. However, there are also cases in which the method fails by either failing to filter OOD samples or by being too conservative and filtering perfectly accurate predictions.
% ==============
% DISCUSSION
% ==============
% Generally I want to use this section to discuss the findings so far as well as build a connection to the methods in the literature: We are more or less trying to achieve the same thing - the process of generating counterfactuals is just so much different in the symbolic graph processing domain than in the continuous image processing domain.
% ~~~
% Picking up on the existing literature on this topic
\section{Discussion}
Previous work has investigated the intersection of uncertainty estimation and counterfactual explainability predominantly in the context of image classification \cite{delaneyUncertaintyEstimationOutofDistribution2021,schutGeneratingInterpretableCounterfactual2021,antoranGettingCLUEMethod2020}. Schut \etal \cite{schutGeneratingInterpretableCounterfactual2021}, for example, include an ensemble-based uncertainty estimate as a direct objective in the optimization of counterfactual explanations. The authors find this intervention to reduce the likelihood of generating uninformative out-of-distribution samples---or in the words of Freiesleben \cite{freieslebenIntriguingRelationCounterfactual2022} to steer the generation toward true counterfactual explanations rather than mere adversarial examples.

% Relating our work to the literature
In our work, we present a distinct perspective to the existing literature, which differs in two important aspects: We focus on (1) regression tasks in the (2) graph processing domain. In image processing, good counterfactual explanations require the modification of multiple pixel values in a semantically meaningful way. This is framed as a non-trivial optimization objective requiring substantial computational effort. In the graph processing domain, however, the limited number of possible graph modifications makes it computationally feasible to search for counterfactual candidates among a full enumeration of all possible perturbations. Consequently, uncertainty quantification does not have to be included in the generation process itself but instead may serve as a simple filter over this set of possible perturbations. Nevertheless, the objective is the same: to use uncertainty quantification methods to present higher-quality counterfactual explanations to the user.

Another key factor is the difference between classification and regression tasks. While classification enables binary assessments of correct and incorrect predictions, regression operates on a continuous error scale, requiring different metrics to assess the impact of uncertainty estimation---motivating our definitions of the UER-AUC and the counterfactual truthfulness. \\

% Discussing the results of our experiments in general.
\noindent Consistent with existing literature, our work demonstrates that incorporating uncertainty estimation improves the quality of counterfactual explanations. We specifically find that filtering high-uncertainty elements decreases the average error of the remaining set and increases overall \textit{truthfulness}---meaning the explanation's alignment with the underlying ground truth data distribution.

In our experiments, we find no substantial differences in the relative effectiveness of UQ interventions between three common graph neural network architectures (GCN, GATv2, GIN). Regarding the choice of the uncertainty estimation method, we find trust scores \cite{delaneyUncertaintyEstimationOutofDistribution2021,schutGeneratingInterpretableCounterfactual2021} to be ill-suited to graph processing applications, most likely due to the unavailability of suitable distance metrics. We furthermore come to similar conclusions as previous authors \cite{scaliaEvaluatingScalableUncertainty2020,hirschfeldUncertaintyQuantificationUsing2020,buskCalibratedUncertaintyMolecular2021} in that the simple application of model ensembles already proves relatively effective. While we find a combination between ensembles and mean variance estimation to be slightly beneficial on IID data, there seems to be no substantial difference in OOD test scenarios. 

However, in this context, it is still important to mention the remaining limitations of this approach grounded in the non-perfect correlation of the prediction errors and estimated uncertainties. While quantitative results show that uncertainty-based filtering has a higher relative likelihood to remove truly high-error samples, qualitative results indicate it still occasionally fails to detect some high-error samples and mistakenly filters valid elements. Depending on the severity of these failure cases, it will largely depend on the concrete application whether the increased truthfulness reasonably justifies the loss of some valid explanations.

\section{Conclusion}

% recap of the motivation
Counterfactual explanations can deepen the understanding of a complex model's predictive behavior by illustrating which kinds of local perturbations a model is especially sensitive to. In the scientific domain of chemistry and material science, these explanations are often not only desirable to understand the model's behavior but by extension to understand the structure-property relationships of the underlying data itself. To use counterfactuals to gain insights about the underlying data, the explanations \textit{truthfully} must reflect the properties thereof.

% summary of the method
In this work, we explore the potential of uncertainty estimation to increase the overall truthfulness of a set of counterfactuals by filtering those elements with particularly high predicted uncertainty. We conduct extensive computational experiments with different methods to investigate the error-reduction potential of various uncertainty estimation methods in different settings. We find that model ensembles provide strong uncertainty estimates in out-of-distribution test scenarios, while a combination of ensembles and mean variance provide the highest error reduction potential for in-distribution settings.

% outlook
Based on these initial results, we conclude that uncertainty estimation presents a promising opportunity to increase the truthfulness of explanations---to make sure explanations not only represent the model's behavior but the properties of the underlying data as well. An interesting direction for future research will be to see if uncertainty estimation can be employed equally beneficially to different explanation modalities, such as local attributional and global concept-based explanations.

\begin{credits}
\subsubsection{\ackname} This work was supported by funding from the pilot program Core-Informatics of the Helmholtz Association (HGF).
\subsubsection{\discintname}
The authors have no competing interests to declare that are
relevant to the content of this article.
\end{credits}

% REFERENCES
% ==========

%\setcitestyle{numbers}
\bibliographystyle{plain}
\bibliography{main.bib}
%\thebibliography

\clearpage

\end{document}